\documentclass[letterpaper, 10 pt, conference]{ieeeconf}  

\IEEEoverridecommandlockouts                              

\usepackage{graphicx} 
\usepackage{amsmath} 
\usepackage{amssymb}  

\usepackage{ifluatex,ifxetex} 
\ifluatex
  \usepackage{fontspec} 
\else\ifxetex
  \usepackage{fontspec} 
\else 
  \usepackage[T1]{fontenc}
  \usepackage[utf8]{inputenc} 
\fi\fi

\usepackage[sorting = none, backend = bibtex, style=numeric-comp]{biblatex}
\usepackage{tabularx}

\newcommand{\dd}{\mathop{}\!\mathrm{d}}

\usepackage{color}

\usepackage[normalem]{ulem}
\newcommand{\edit}[2]{#2}

\title{\LARGE \bf
Rollbot: a Spherical Robot Driven by a Single Actuator
}

\author{Jingxian Wang$^{1}$ and Michael Rubenstein$^{1}$
\thanks{$^{1}$Jingxian Wang and Michael Rubenstein are with the Center for Robotics and Biosystems,
        Northwestern University, Evanston, IL 60601, USA
        {\tt\small jingxianwang2026@u.northwestern.edu, rubenstein@northwestern.edu}}%
}

\begin{document}

\maketitle
\thispagestyle{empty}
\pagestyle{empty}

\begin{abstract}

\edit{}{Spherical robots typically require at least two actuators to achieve controlled 2D planar motion.} Here we present Rollbot, the first spherical robot capable of controllably maneuvering on a 2D plane with a single actuator, \edit{}{challenging this assumption}. Rollbot rolls on the ground in a circular pattern and controls its motion by changing the trajectory's curvature by accelerating and decelerating its single motor and the attached mass \edit{}{according to our derived quasi-stable state dynamics and control laws}. We present the theoretical analysis, design, and control of Rollbot, and demonstrate its ability to move in a controllable circular pattern and follow waypoints\edit{}{, validating the efficacy of the proposed theoretical framework}.


\end{abstract}

\section{Introduction}

Underactuated robots have attracted researchers' attention in recent years due to their advantages such as energy savings, material savings, and space savings \cite{urreview}. The simplicity of underactuated robots is also often valued in swarm robotics \cite{Wang-RSS-24}. While underactuated systems like inverted pendulums \cite{pendulumreview,quadpendulum}, underactuated hands \cite{urhand1,urhand2}, and underactuated wrists \cite{urwrist1,urwrist2} have been extensively studied, few mobile robots have adopted the idea of extreme underactuation.

Currently, only a handful of mobile robots can maneuver in 2D or 3D space with a single actuator. Robots in \cite{zarrouk2012compliance, bernardes2023modelling,zhao2013msu} use compliant mechanisms or one-way bearings to allow a single actuator to achieve multiple functions or modes of operation, effectively `multiplexing' the actuator to move in space. Mugatu in \cite{kyle2023simplest} controls its movement by varying the step sizes of its left and right legs. Robots in \cite{knizhnik2021thrust,saloutos2019spinbot,wang2022pcbot,ito2019autonomous} and drones in \cite{zhang2016controllable,curtis2023autonomous,Wang-RSS-24} move by rotating their bodies and accelerating their motors at appropriate times. Among these works, only \cite{ito2019autonomous} proposes a wheeled robot, which is also the only ground robot that does not move in discrete steps.

Another key inspiration for our work is spherical robots \cite{diouf2024spherical}, which are studied for their structural simplicity and the inherent protective nature of their spherical shells that shields the internal mechanisms and electronics from hostile environments. Since the pioneering spherical robot \cite{halme1996motion}, three major types of driven mechanisms for spherical robots have been proposed \cite{diouf2024spherical}. These include the most common barycentric type \cite{mojabi2002introducing,chen2016design}, which moves masses attached to pendulums or tracks on the robot; the conservation of angular momentum (CoAM) type \cite{bhattacharya2000spherical}, which uses reaction wheels; and the shell deformation type \cite{wait2010self}, which changes the shape of the shell. Interestingly, to the authors’ knowledge, only one spherical robot with a single actuator has been reported \cite{spigler2024intuitive}, and it has not been validated in the real world. Even the authors of the review paper \cite{diouf2024spherical} claimed that ``For a robot to be able to move in more than one direction, at least two (control) DoFs are needed.''
Rollbot decisively challenges this assumption, demonstrating for the first time that a spherical robot can achieve precise and controllable 2D motion in the real world using a single actuator.

Furthermore, Rollbot's design and movement algorithm can be used as a fail-safe for regular barycentric type spherical robots. If properly designed, regular spherical robots can downgrade to the configuration of the proposed robot when only one motor remains active, enabling continued operation despite the limited functionality. This feature is especially helpful as spherical robots typically work in hostile environments like on the Moon or on Mars, where a motor failure could end the mission.

\begin{figure}[!t]
\centering
\includegraphics[width=\linewidth]{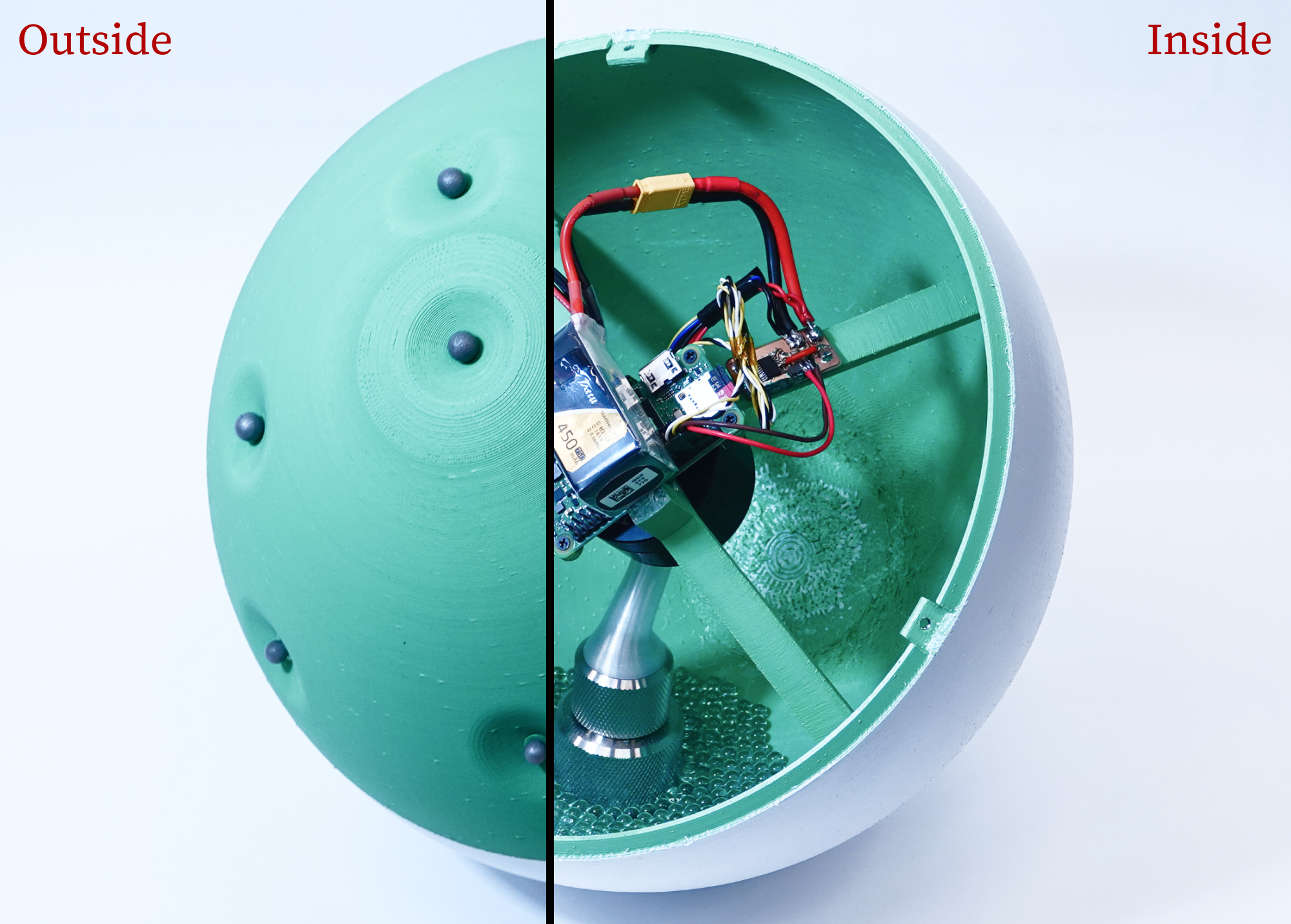}
\vspace{-7mm}
\caption{Photo of inside and outside of Rollbot. Rollbot has an outer diameter of $24\,$cm and weighs $1.2\,$kg.}
\vspace{-7mm}
\label{fig:rollbotphoto}
\end{figure}

\begin{figure}[!t]
\centering
\includegraphics[width=\linewidth]{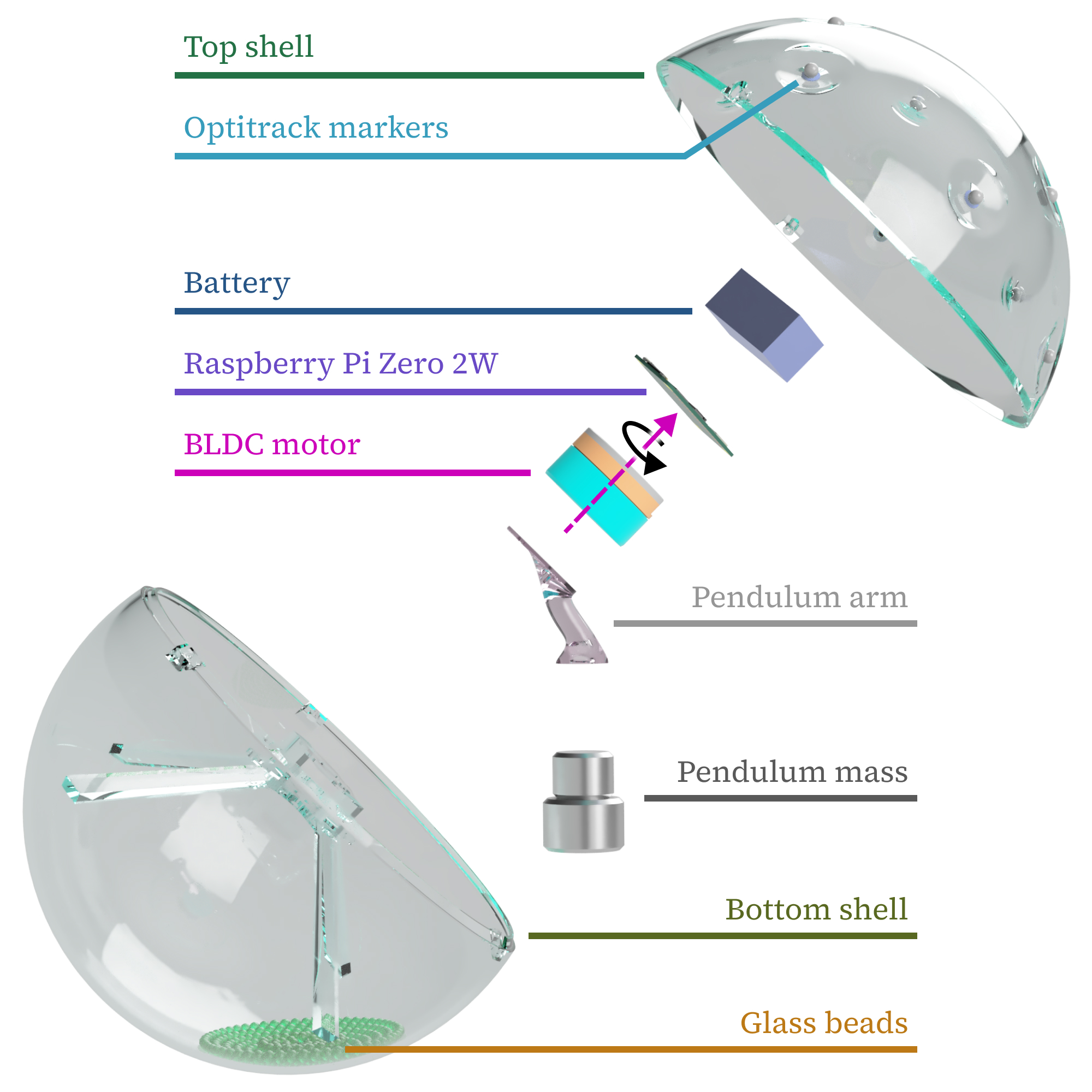}\\
\vspace{-1.5mm}
\caption{An exploded view of Rollbot. The only actuator is the BLDC motor at the center, which can rotate around the purple axis.}
\vspace{-1mm}
\label{fig:exploded}
\end{figure}

The lack of research on single-actuator spherical robots, despite their potential benefits such as structural simplicity, cost savings, and applications in actuation failure scenarios and swarm robotics, motivates us to develop Rollbot, a barycentric spherical robot driven by a single motor. Rollbot is presented as a proof-of-concept platform demonstrating that precise planar locomotion of a spherical robot can be achieved using a single actuator. While the current implementation prioritizes concept validation over performance, future improvements in parameter selection, structural design, and advanced control can enable single actuator spherical robots to achieve performance comparable to that of more complex conventional spherical robots.

The notation and symbols used in this study are presented in Table. \ref{tab:nomenclature}.

\begin{table}[t]
    \centering
    \begin{tabularx}{0.45\textwidth}{ |c|X| }
        \hline
        \textbf{Symbol} & \textbf{Definition} \\
        \hline
        $\vec{a}$ & Vector quantity \\
        $\hat{a}$ & Unit vector in the direction of $\vec{a}$ \\
        $\hat{x}$ & Unit vector in the positive $x$-axis direction \\
        $[\vec{a}]$ & Skew-symmetric matrix representation of $\vec{a}$ \\
        $\mathbf{T}$ & Matrix quantity \\
        $\mathbf{I}_3$ & $3 \times 3$ identity matrix \\
        $\dot{a}$ & Time derivative of $a$ \\
        $\vec{a}\vec{c}$ & Outer product of vectors $\vec{a}$ and $\vec{c}$ \\
        $a^p$ & Perturbed value of $a$ \\
        $\delta a$ & Perturbation of $a$, defined as $\delta a = a^p - a$ \\
        \hline
    \end{tabularx}
    \caption{Notation and conventions used in this paper.}
    \label{tab:nomenclature}
    \vspace{-8mm}
\end{table}

\section{Robot Design}\label{sec:design}

Rollbot, as shown in Fig. \ref{fig:rollbotphoto} and \ref{fig:exploded}, is designed to test the feasibility of achieving 2D motion using only a single motor actuator. Rollbot is composed of a spherical shell and a pendulum connected to the shell through a motor. As the motor rotates the pendulum, the shell rolls on the ground, and we can control Rollbot's motion by changing the motor's rotation speed. For conciseness, we will refer to the spherical shell as `the shell', the pendulum mass as `the mass', and the motor's spinning speed as the `driving speed' in subsequent discussions.

\subsection{Dynamics of Rollbot}\label{sec:dynamics}

\begin{figure}[htbp]
\centering
\includegraphics[width=0.9\linewidth]{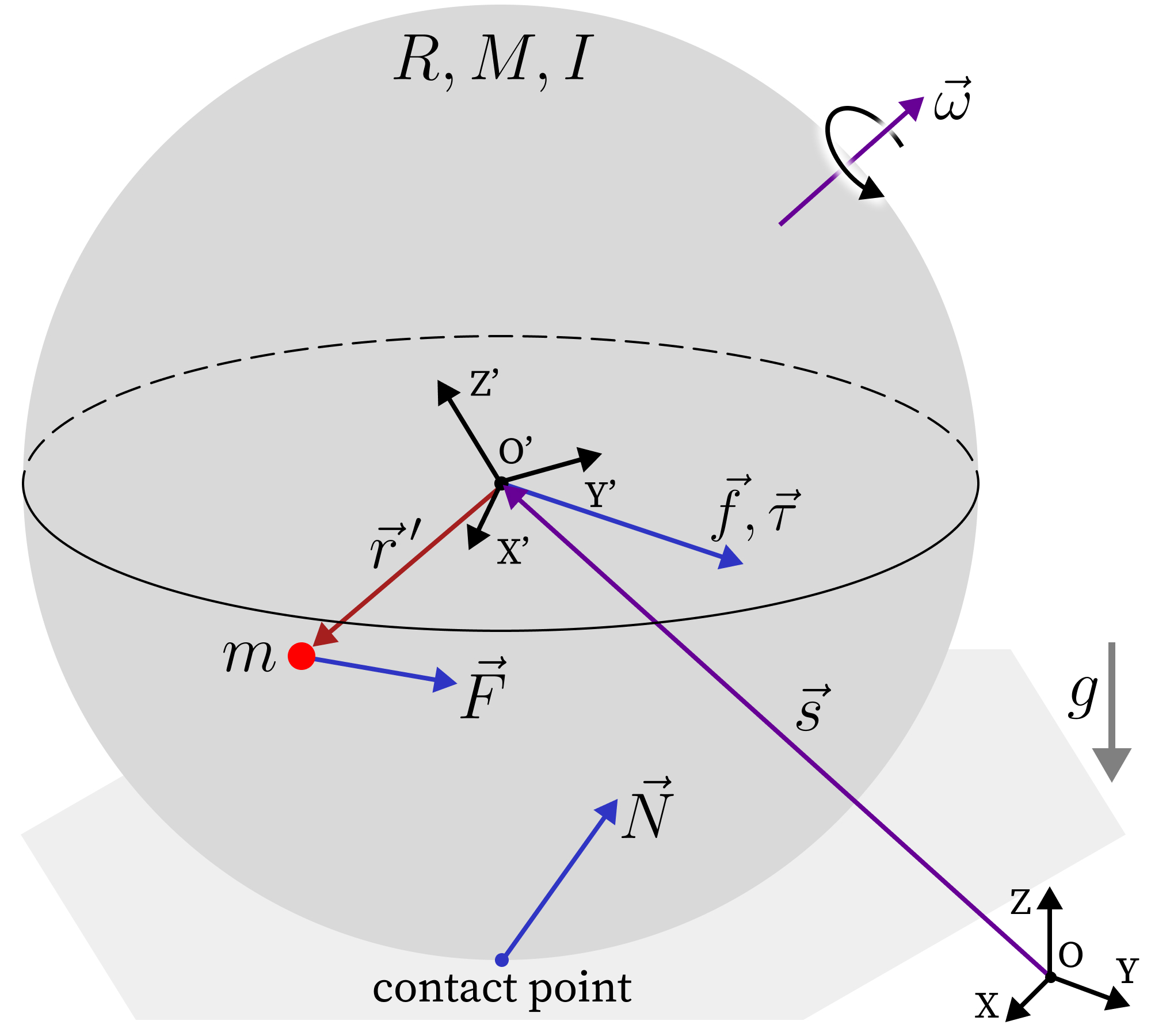}\\
\vspace{-1.5mm}
\caption{Illustration of relevant quantities. O-XYZ is the ground reference frame $G$, O'-X'Y'Z' is the shell's body reference frame $B$. $\vec{s}$ is the position of the center of the shell in $G$, $\vec{\omega}$ is the angular velocity of the shell in $G$, and $\vec{r}\,'$ is the position of the mass in $B$. $R,M,I$ are the radius, mass, and the moment of inertia of the shell respectively, and $m$ is the mass of the point mass. $\vec{f},\vec{\tau}$ are the externally applied force and torque on the shell with respect to $O'$, $\vec{N}$ is the sum of the normal and friction force the ground applies to the shell, and $\vec{F}$ is the force the shell applies to the point mass.}
\vspace{-5mm}
\label{fig:force}
\end{figure}

Rollbot can be considered as a spherical robot driven by an internal mass moving along a circular trajectory in the shell's reference frame. In this section, we will derive the dynamics of a general spherical robot driven by the movement of internal point masses and apply the results to Rollbot. We will start with a simple case where a spherically symmetric shell is driven by one internal point mass, as illustrated in Fig. \ref{fig:force}. We also assume that the robot maintains contact with the ground and does not slip.

The vector from the center of shell to the point mass $O'm$ in ground frame, $\vec{r}$, can be described using the rotation matrix $\textbf{T}$ from body frame to ground frame as

\vspace{-2mm}
\begin{equation}
    \vec{r} = \textbf{T}\cdot \vec{r}\,'
    \label{eqn:trans}
\end{equation}
\vspace{-3mm}

\noindent We can take its first and second derivatives and obtain

\vspace{-2mm}
\begin{align}
    \dot{\vec{r}} =& \vec{\omega}\times(\mathbf{T}\cdot\vec{r}\,')+\mathbf{T}\cdot\dot{\vec{r}}\,' \label{eqn:rdot}\\
    \ddot{\vec{r}} =& \dot{\vec{\omega}}\times(\mathbf{T}\cdot\vec{r}\,')\nonumber\\
    &+(\vec{\omega}\times(\vec{\omega}\times(\mathbf{T}\cdot\vec{r}\,'))+2\vec{\omega}\times(\mathbf{T}\cdot\dot{\vec{r}}\,')+\mathbf{T}\cdot\ddot{\vec{r}}\,')\nonumber\\
     =&\dot{\vec{\omega}}\times(\mathbf{T}\cdot\vec{r}\,')+\vec{h}(\vec{\omega},\mathbf{T},\vec{r}\,'(t)) \label{eqn:rddot}
\end{align}
\vspace{-3mm}

\noindent Notice that the term $\vec{h}$ is the acceleration of point mass in the non-accelerating co-rotating frame, including the centrifugal, Coriolis, and relative acceleration terms but does not include acceleration of the shell.

Newton's second law and the angular momentum theorem yield the equations of motion of the point mass and the shell.

\vspace{-2mm}
\begin{align}
    m(\ddot{\vec{s}}+\ddot{\vec{r}})&=\vec{F}-mg\hat{z} \label{eqn:m}\\
    M\ddot{\vec{s}}&=-\vec{F}+\vec{N}-Mg\hat{z}+\vec{f} \label{eqn:M}\\
    I\dot{\vec{\omega}}&=-(\vec{r}-\vec{s})\times \vec{F}+(-R\hat{z})\times \vec{N} + \vec{\tau} \label{eqn:I}
\end{align}
\vspace{-3mm}

\noindent We can eliminate $\vec{F}$ and $\vec{N}$ by combining the three equations

\vspace{-2mm}
\begin{equation}
    \vec{u}\times(\text{Eqn.} \ref{eqn:m})+R\hat{z}\times(\text{Eqn.} \ref{eqn:M})+(\text{Eqn.} \ref{eqn:I})
\end{equation}
\vspace{-3mm}

\noindent where $\vec{u}=R\hat{z}+\vec{r}$ is the vector from the contact point to the point mass and obtain

\vspace{-2mm}
\begin{equation}
    \begin{aligned}
    \vec{u}\times&(m(\ddot{\vec{s}}+\ddot{\vec{r}}\,))+(R\hat{z})\times(M\ddot{\vec{s}}\,)+I\dot{\vec{\omega}}\\
    &=\vec{u}\times(-mg\hat{z})+R\hat{z}\times\vec{f}+\vec{\tau}
    \end{aligned}
    \label{eqn:motionpre1}
\end{equation}
\vspace{-2mm}

\noindent Then we can apply the no slipping constraint

\vspace{-2mm}
\begin{equation}
    \dot{\vec{s}}=\vec{\omega}\times(R\hat{z})
    \label{eqn:noslip}
\end{equation}
\vspace{-3mm}

\noindent to (\ref{eqn:motionpre1}) and substitute $\ddot{\vec{r}}$ using (\ref{eqn:rddot}) to obtain the complete equation of motion of the shell as

\vspace{-2mm}
\begin{equation}
    \begin{aligned}
    &(m(\vec{u}\cdot\vec{u}\,\mathbf{I}_3-\vec{u}\vec{u})+M R^2(\mathbf{I}_3-\hat{z}\hat{z})+I\,\mathbf{I}_3)\cdot\dot{\vec{\omega}}\\
    &=m \vec{u}\times(-\vec{h}(\vec{\omega},\mathbf{T},\vec{r}\,'(t))-g\hat{z})+R\hat{z}\times\vec{f}+\vec{\tau}
    \end{aligned}
    \label{eqn:motion}
\end{equation}
\vspace{-2mm}

The physical meaning of (\ref{eqn:motion}) is clear: the left hand side is the moment of inertia matrix of the whole robot with respect to the contact point applied on the angular acceleration. The right hand side is the torque applied on the robot with respect to the contact point contributed by the point mass's inertia in the non-accelerating co-rotating frame, the point mass's gravity, and the external force and torque applied on the shell.

We can also write down the sum of supporting and friction force

\vspace{-2mm}
\begin{equation}
    \vec{N}=(M+m)\ddot{\vec{s}}+m\ddot{\vec{r}}+(M+m)g\hat{z}-\vec{f}
\end{equation}
\vspace{-3mm}

\noindent The condition for the robot to not jump is the supporting force being positive, i.e.

\vspace{-2mm}
\begin{equation}
    \vec{N}\cdot\hat{z}>0
\end{equation}
\vspace{-3mm}

\noindent The condition for the robot to not slide is the friction force being no larger than the friction coefficient, $\mu$, times the normal force, i.e.

\vspace{-2mm}
\begin{equation}
    \frac{\vec{N}\cdot\hat{z}}{\|\vec{N}\|}\geq\frac{1}{\sqrt{1+\mu^2}}
\end{equation}
\vspace{-2mm}

The derived equation of motion can be extended to handle more general cases. Adding a summation over all terms related to the point mass extends the formulation to multiple point masses

\vspace{-2mm}
\begin{equation}
    \begin{aligned}
    &\left(\sum\limits_q m_q(\vec{u}_q\cdot\vec{u}_q\,\mathbf{I}_3-\vec{u}_q\vec{u}_q)+M R^2(\mathbf{I}_3-\hat{z}\hat{z})+I\,\mathbf{I}_3\right)\cdot\dot{\vec{\omega}}\\
    &=\sum\limits_q m_q \vec{u}_q\times(-\vec{h}(\vec{\omega},\mathbf{T},\vec{r}\,'_q(t))-g\hat{z})+R\hat{z}\times\vec{f}+\vec{\tau}
    \end{aligned}
    \label{eqn:generalmotion}
\end{equation}
\vspace{-2mm}

\noindent where $m_q$ is the mass of $q$th point mass, $\vec{r}\,'_q$ is the position of $q$th point mass in $B$, and $\vec{u}_q$ is the vector from the contact point to $q$th point mass. Furthermore, by distributing extra point masses fixed to the shell or other point masses, (\ref{eqn:generalmotion}) can also handle cases involving non-symmetrical mass distribution of the shell or general solid bodies moving inside the robot.

The result we obtained is equivalent to that proposed in \cite{putkaradze2018dynamics}, but the derivation is more straightforward and provides clearer physical insights.

When applying the model to Rollbot's case, a good approximation is to use (\ref{eqn:motion}) and let 

\vspace{-2mm}
\begin{align}
    \vec{r}\,'=r_0 (\sin\phi \cos\theta(t) \, \hat{x}&+\sin\phi \sin\theta(t) \, \hat{y}-\cos\phi \, \hat{z})\\
    \vec{\tau}&=-k_0 \vec{\omega} \label{eqn:damp}\\
    \vec{f}&=\mathbf{0}\\
    M&=M_s+m_b
\end{align}
\vspace{-3mm}

\noindent where $\theta(t)$ is the rotating angle of the pendulum mass around $Z'$ axis, and $k_0$ is an experimentally determined damping constant. Such simplification is valid when:

\begin{itemize}
    \item The shell has a moment of inertia matrix in the form of $I\,\mathbf{I}_3$.
    \item The pendulum mass's geometric size is small compared to both its distance from the center $|\vec{r}\,'|$ and the shell's radius $R$, so we can simplify the pendulum mass to a point mass.
    \item The damping on the shell's motion is small and has small impact on the behavior of Rollbot. 
\end{itemize}

\noindent and we will show that our system satisfies these requirements in section \ref{sec:hardware}.

\subsection{Quasi-stable State of Rollbot}\label{sec:static}

\begin{figure}[htbp]
\centering
\vspace{-2mm}
\includegraphics[width=.9\linewidth]{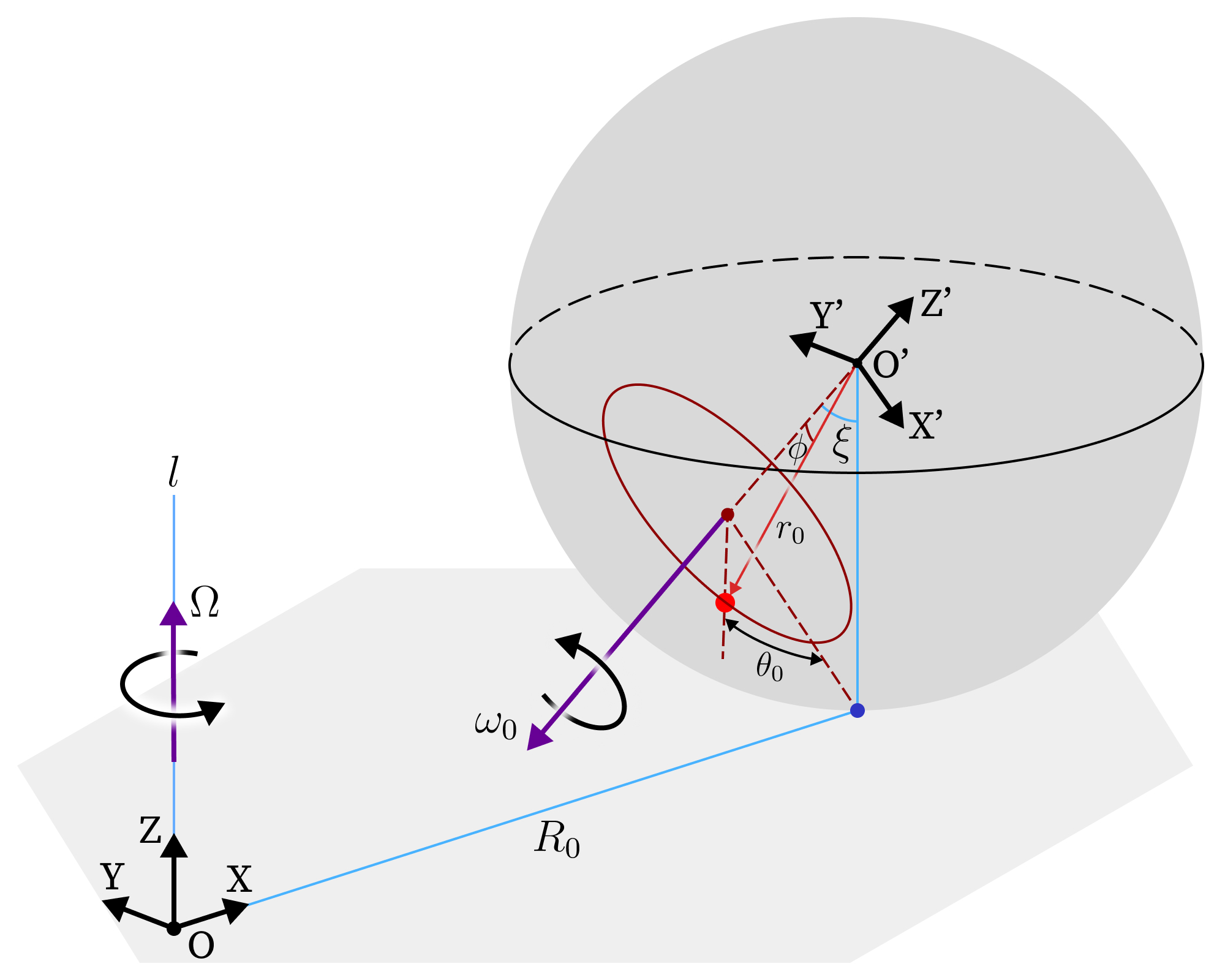}\\
\vspace{-1.5mm}
\caption{Illustration of the quasi-stable state at $t=0$. O-XYZ is the stationary ground reference frame $G$ with its origin at the intersection of the revolving axis $l$ and the ground plane, and X axis pointing toward the contact point. O'-X'Y'Z' is the shell's body reference frame $B$ at this moment. Without loss of generality, we place Y' axis parallel to Y axis and Z' axis aligned with the rotation axis of the motor and on X-Z plane, and the angle between Z and Z' is $\xi$. The shell revolves around $l$ at angular velocity $\Omega \hat{z}$, while also spinning at angular velocity $\vec{\omega}_s = - \omega_0 \hat{z}'$ in the quasi-stable state. The shell's revolving radius is $R_0$.}
\vspace{-2mm}
\label{fig:quasi-stable}
\end{figure}

In our experiments, we slowly accelerate and decelerate the pendulum mass, keeping Rollbot close to the quasi-stable state where the driving speed $\theta'(t)=\omega_0$ is constant. In the quasi-stable state, Rollbot revolves around a vertical axis $l$. In the frame revolving together with the Rollbot, its state remains constant, as shown in Fig. \ref{fig:quasi-stable}.

Because the entire system revolves around $l$ at constant angular velocity $\Omega$, we have 

\vspace{-2mm}
\begin{align}
    \vec{u}(t) &= \mathbf{T}_r(t) \cdot \vec{u}(0) \label{eqn:ut}\\
    \vec{\omega}(t) &= \mathbf{T}_r(t) \cdot \vec{\omega}(0) \label{eqn:ot}
\end{align}
\vspace{-2mm}

\noindent where $\mathbf{T}_r(t)$ is the rotation matrix defined as

\vspace{-2mm}
\begin{equation}
    \mathbf{T}_r(t)=\begin{bmatrix}
        \cos(\Omega t) & -\sin(\Omega t) & 0 \\
        \sin(\Omega t) & \cos(\Omega t) & 0 \\
        0 & 0 & 1
    \end{bmatrix}
    \label{eqn:Tr}
\end{equation}
\vspace{-2mm}

\noindent Using this property, we can then express (\ref{eqn:motion}) using only variables at $t=0$, and subsequently reduce it into three equations about $\Omega$, $\theta_0$, and $\xi$.

\begin{figure}[htbp]
\centering
\vspace{-2mm}
\includegraphics[width=.95\linewidth]{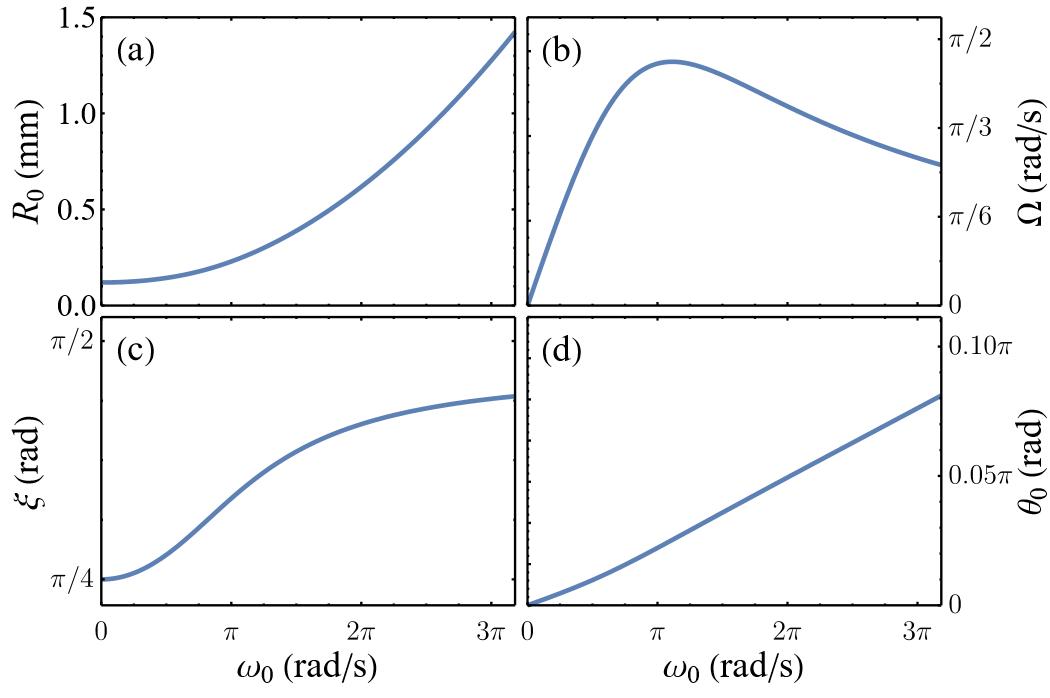}\\
\vspace{-3mm}
\caption{Plots of key parameters of quasi-stable state versus driving speed $\omega_0$. (a) to (d) show the trends of revolving radius $R_0$, revolving angular velocity $\Omega$, tilting angle of rotating axis of the motor $\xi$ and rotating angle of pendulum mass $\theta_0$ respectively. \edit{}{These results are calculated using the specific physical parameters of the physical Rollbot prototype as detailed in Table \ref{tab:params}.}}
\vspace{-1mm}
\label{fig:func}
\end{figure}

\vspace{-2mm}
\begin{equation}
\begin{aligned}
    m\vec{u}&\times(-g\vec{z}+\Omega^2(R_0 \hat{x}+\vec{u}-(\vec{u}\cdot\hat{z})\hat{z})+M\Omega^2R_0 R \hat{y}\\
    &-k_0(-\omega_0\hat{z}'+\Omega \hat{z})+I\Omega\omega_0\sin\xi \hat{y}=\mathbf{0}
\end{aligned}
\label{eqn:quasi}
\end{equation}
\vspace{-2mm}

\noindent where $R_0=\omega_0 R \sin\xi / \Omega$ and $\hat{z}'=\sin\xi \hat{x}+\cos\xi\hat{z}$. The quasi-stable state of Rollbot at different driving speed $\omega_0$ can then be obtained by numerically solving (\ref{eqn:quasi}). The solution with Rollbot’s actual physical parameters is shown in Fig. \ref{fig:func}.

Fig. \ref{fig:func} shows that the revolving radius $R_0$ starts at $R\tan\phi$ when $\omega_0=0$, and gradually increases with $\omega_0$, suggesting that we can control the trajectory of Rollbot by changing the driving speed. $\xi$ starts at $\phi$ when $\omega_0=0$ and gradually approaches $\pi/2\,\text{rad/s}$, meaning that Rollbot moves in a nearly straight line when rolling at high speeds. $\theta_0$ remains close to $0$, meaning that the pendulum mass stays near the bottom of the shell when driving speed is smaller than $3\pi\,\text{rad/s}$.

Furthermore, we also discovered that as long as damping coefficient $k_0$ is not excessively large, it has little impact on the results. This means that we can make assumption (\ref{eqn:damp}) for convenience of computation without significantly affecting the results.

Aside from the state shown above, there is also another set of `fast revolving state', where $\theta_0$ is close to $\pi$ and Rollbot revolves in the opposite direction at a much smaller radius. However, because Rollbot moves quasi-stably in our experiments, this set of states will not occur.

\subsection{Perturbation around Quasi-stable State} \label{sec:perturb}

Perturbation around the quasi-stable state tells us how stable the quasi-stable state is and the response of Rollbot if we deviate from the state, either during acceleration and deceleration or due to external disturbances. We will only consider the first order perturbation in the following discussion.

We can apply a perturbation to the rotation matrix $\mathbf{T}$ and have

\vspace{-2mm}
\begin{equation}
    \mathbf{T}^p(t)=(1+[\mathbf{T}_r(t)\cdot\vec{\alpha}(t)])\cdot\mathbf{T}(t)
\end{equation}
\vspace{-2mm}

\noindent where $\vec{\alpha}$ is a small perturbation vector. According to the definition, we have

\vspace{-2mm}
\begin{align}
    \delta\vec{u} = \delta\vec{r} &= (\mathbf{T}_r\cdot\vec{\alpha})\times\vec{r} \nonumber\\
    &=\mathbf{T}_r\cdot(\vec{\alpha}\times\vec{r}(0)) \label{eqn:up}\\
    \delta\vec{\omega} &= (\mathbf{T}_r\cdot\vec{\alpha})\times\vec{\omega}+d_t(\mathbf{T}_r\cdot\vec{\alpha}) \nonumber\\
    &= \mathbf{T}_r\cdot(\vec{\alpha}\times\vec{\omega}_s(0)+\dot{\vec{\alpha}}) \label{eqn:op}
\end{align}
\vspace{-2mm}

\noindent Put (\ref{eqn:up}) and (\ref{eqn:op}) into (\ref{eqn:noslip}) and we have

\vspace{-2mm}
\begin{equation}
    \delta\dot{\vec{s}} = \mathbf{T}_r\cdot((\vec{\alpha}\times\vec{\omega}_s(0)+\dot{\vec{\alpha}})\times(R\hat{z}))
    \label{eqn:sp}
\end{equation}
\vspace{-3mm}

\noindent Take perturbation of (\ref{eqn:motionpre1}), left multiply by $\mathbf{T}_r^{-1}$ and use (\ref{eqn:sp}) then we have

\vspace{-3mm}
\begin{equation}
    \begin{aligned}
    &m(\vec{\alpha}\times\vec{r}(0))\times(\ddot{\vec{u}}(0)+\ddot{\vec{s}}(0))\\
    &+m\vec{u}(0)\times\Omega^2(\vec{\alpha}\times\vec{r}(0)-((\vec{\alpha}\times\vec{r}(0))\cdot\hat{z})\hat{z})\\
    &+m\vec{u}(0)\times(2\Omega\hat{z}\times(\dot{\vec{\alpha}}\times\vec{r}(0))+\ddot{\vec{\alpha}}\times\vec{r}(0))\\
    &+(m\vec{u}(0)+MR\hat{z})\times((\dot{\vec{\alpha}}\times\vec{\omega}_s(0)+\ddot{\vec{\alpha}})\times(R\hat{z}))\\
    &+I(\dot{\vec{\alpha}}\times\vec{\omega}_s(0)+\ddot{\vec{\alpha}})\\
    &+mg(\vec{\alpha}\times\vec{r}(0))\times\hat{z}\\
    &+k_0(\vec{\alpha}\times\vec{\omega}_s(0)+\dot{\vec{\alpha}})=0
    \end{aligned}
    \label{eqn:perturb}
\end{equation}
\vspace{-2mm}


\noindent Notice that this is a second order ordinary differential equation group of $\vec{\alpha}(t)$ and the parameters are not dependent on time, we can plug in the values obtained in section \ref{sec:static} and obtain an equation in the following form

\vspace{-2mm}
\begin{equation}
    \begin{pmatrix}
        \dot{\vec{\alpha}} \\
        \ddot{\vec{\alpha}}
    \end{pmatrix}
    =
    \begin{pmatrix}
        0 & \mathbf{I}_3 \\
        \mathbf{A}_{3\times 3} & \mathbf{B}_{3\times 3}
    \end{pmatrix}
    \cdot
    \begin{pmatrix}
        \vec{\alpha} \\
        \dot{\vec{\alpha}}
    \end{pmatrix}
    \label{eqn:ode}
\end{equation}
\vspace{-2mm}

\noindent The recovery behavior of Rollbot under perturbation is characterized by eigenvalues of the matrix in (\ref{eqn:ode}).

In addition, notice that the system will be identical if we make a perturbation and let Rollbot revolve around $l$ by a certain angle, we should remove the eigenvalue and eigenvector corresponding to this mode of movement when analyzing the stability. Finally, we can conclude that if all but one eigenvalues have real part smaller than 0, then Rollbot is stable under perturbation and the characteristic time of recovery $\tau$ is determined by the eigenvalue $\lambda_2$ with second smallest real part.

\vspace{-2mm}
\begin{equation}
    \tau \approx -\frac{1}{Re(\lambda_2)}
\end{equation}
\vspace{-2mm}

\begin{figure}[htbp]
\centering
\vspace{-2mm}
\includegraphics[width=.85\linewidth]{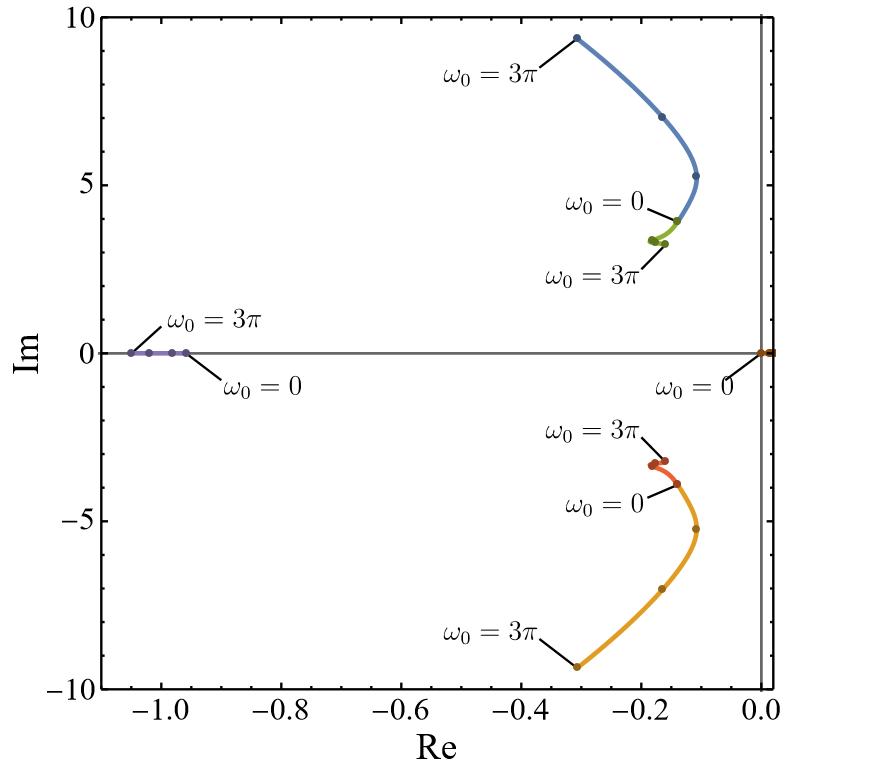}\\
\vspace{-3mm}
\caption{Plot of eigenvalues' trajectory for $\omega_0\in[0,3\pi]$. \edit{}{Only one eigenvalue corresponding to revolution symmetry around $l$ has positive real part, meaning such motion is stable. The second largest real part of all eigenvalues, $Re(\lambda_2)$, is within $[-0.1,-0.2]$, meaning the characteristic recovery time from transient behavior is around $\tau=7\,\text{s}$. These results are calculated using the specific physical parameters of the physical Rollbot prototype as detailed in Table \ref{tab:params}.}}
\vspace{-1mm}
\label{fig:eig}
\end{figure}

Plot of eigenvalues' distribution under different $\omega_0$ with Rollbot’s actual physical parameters is shown in Fig. \ref{fig:eig}, showing that with our choice of physical parameters given in Table. \ref{tab:params}, Rollbot is able to recover from perturbations with a characteristic time of about $7\,$s. This is consistent with the result in direct numerical simulation as well as our experiments.

As far as the authors know, this is the first time quasi-stable state and perturbation behavior are analyzed for spherical robots.

\subsection{Hardware Design}\label{sec:hardware}

When designing the hardware, we want it to satisfy the following criterion:

\begin{itemize}
    \item Rollbot should be as close to our theoretical model as possible, i.e. satisfying the criteria given in section \ref{sec:dynamics}.
    \item Rollbot should be close to quasi-stable state even when accelerating/decelerating.
    \item Rollbot's trajectory should change significantly as we accelerate/decelerate the motor.
\end{itemize}

The first criteria require us to design the Rollbot shell's center of mass near the shell's geometric center, moment of inertia in the form of $I_0 \mathbf{I}_3$, and pendulum mass small enough to be considered as a point mass, and the second requires Rollbot to recover quickly from transient state, and the third requires revolving radius $R_0$ to change significantly when we change the driving speed $\omega_0$. 

We designed the diameter of Rollbot's shell to be $24\,$cm, which is a few times larger than the scale of the battery and the motor, and concentrated the mass at the spherical surface of the shell. This makes sure that the moment of inertia is dominated by the surface of the shell instead of its internal structures, keeping the moment of inertia in the form of $I_0 \mathbf{I}_3$. Furthermore, we also carefully positioned the motor and the battery to keep the center of mass near the the shell's geometric center. The pendulum mass is made of stainless steel and is small compared to the distance between it and the center of the shell as well as the size of the shell. We put $m_b=40\,$g of glass beads in the shell so the collision between beads can introduce energy dissipation and create damping\edit{}{, making sure that the Rollbot can quickly recover from transient behaviors}. Because the total mass of the beads is much smaller than the pendulum mass or the shell, and beads remain near the bottom of the shell, we can simplify the beads' effect to an addition of mass to the shell plus a linear damping given by (\ref{eqn:damp}).

With the analysis in section \ref{sec:perturb}, we found out that the system is more stable when $r_0$, $m$, $\phi$, and $k_0$ are larger. As such, we placed the pendulum mass as close to the shell as possible, and introduced the beads inside the shell. Experimentally, comparing to the case where no beads are added, Rollbot's quasi-stable dynamics is almost the same while exhibiting much less oscillation and much faster recovery after acceleration and deceleration.

With the analysis in section \ref{sec:static}, we can notice that the revolving radius $R_0$ changes more significantly with $\omega_0$ when $\phi$ and $m$ are smaller. This factor is conflicting with the stability factor discussed in the previous paragraph. We chose a balanced set of parameters shown in Table. \ref{tab:params}, which allows Rollbot to vary its revolving radius from $0.12\,$m to $1.28\,$m when $\omega_0\in[0,3\pi]\,\text{rad/s}$ and recover from transients within a few seconds.


\begin{table}[h]
    \centering
    \begin{tabularx}{.49\textwidth}{ |c|l|X| }
        \hline
        \textbf{Quantity} & \textbf{Value} & \textbf{Explanation} \\
        \hline
        $R$ & $0.12\,$m & Outer radius of shell\\
        $M$ & $0.840\,$kg & Mass of shell\\
        $I$ & $0.0053\,\text{kg}\cdot\text{m}^2$ & Moment of inertia of shell\\
        $\phi$ & $45^\circ$ & Angle between the pendulum mass, center of shell and the axis of motor \\
        $r$ & $0.093\,$m & Distance between pendulum mass's center of mass and center of shell \\
        $m$ & $0.306\,$kg & Pendulum mass \\
        $m_b$ & $0.040\,$kg & Total mass of beads \\
        $k_0$ & $(0.4\,\text{s}^{-1})MR^2$ & Friction coefficient\\
        \edit{}{$M_{max}$} & \edit{}{$0.28\,\text{N}\cdot\text{m}$} & \edit{}{Maximum rolling torque, computed using $M_{max}=r\,m\,g$}\\
        \hline
    \end{tabularx}
    \caption{Physical parameters of Rollbot.}
    \label{tab:params}
    \vspace{-8mm}
\end{table}

\subsection{Control Design}\label{sec:control}

\begin{figure}[!t]
\centering
\vspace{-1mm}
\includegraphics[width=.8\linewidth]{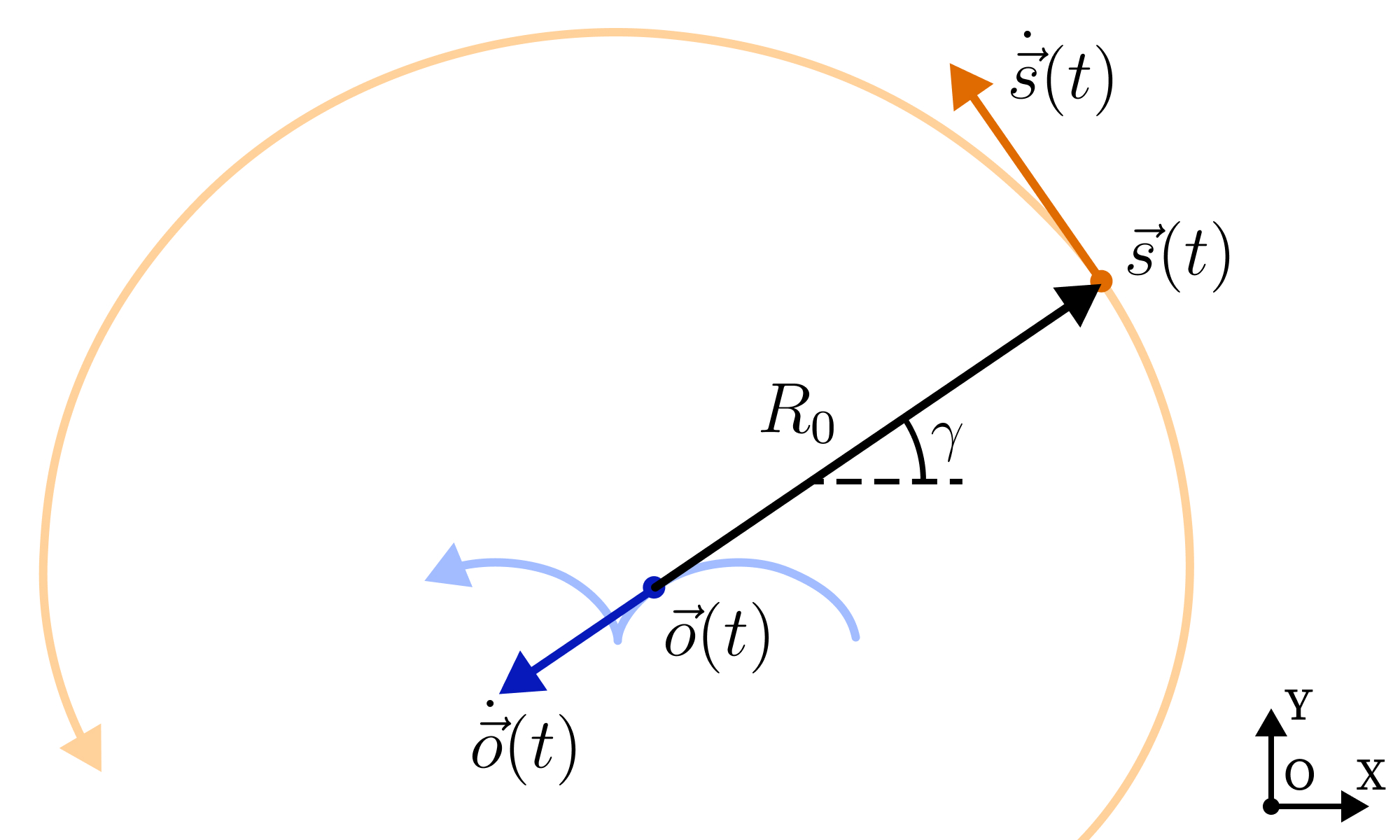}
\caption{Illustration of relevant parameters in the control algorithm. The orange and blue curves are the trajectory of the center of the shell and the corresponding center of curvature respectively. At time $t$, the center of shell is at $\vec{s} \,(t)$, the center of curvature is at $\vec{o}\,(t)$ and the curvature radius is $R_0$. The center of shell is moving at velocity $\dot{\vec{s}}\,(t)$ along the curve perpendicular to $\vec{s}-\vec{o}$, and if Rollbot is accelerating, then the center of curvature will also move along $\vec{s}-\vec{o}$ at a rate of $\dot{\vec{o}}\,(t)$}
\vspace{-6mm}
\label{fig:control}
\end{figure}

We already have a robot that can roll on the ground in circular pattern and theory derived in section \ref{sec:static} tells us that the revolving radius $R_0$ can be modulated by changing the driving speed $\omega_0$. In this section, we will derive a way of controlling Rollbot's 2D motion by changing $R_0$ at different times. \edit{}{An illustration of this process and relevant parameters are shown in Fig.\ref{fig:control}.}

Instead of directly controlling the position $\vec{s}$ of Rollbot, we control the center of curvature $\vec{o}$.

\vspace{-2mm}
\begin{equation}
    \vec{o}=\vec{s}-R_0(\cos\gamma\, \hat{x} + \sin\gamma\,\hat{y})
    \label{eqn:ofunc}
\end{equation}
\vspace{-3mm}

\noindent When Rollbot is rolling with a constant driving speed $\omega_0$, it will move in a circle, and the center of curvature will be stationary. However, if we accelerate the driving speed at a rate $\beta$ at $t=t_0$ when $\vec{s}-\vec{o}$ is at angle $\gamma$ to $\hat{x}$, then the rate of change of curvature will be

\vspace{-2mm}
\begin{equation}
    |\dot{\vec{o}}\,|=v_{R_0}=\left. \frac{\dd  R_0(\omega)}{\dd \omega}\right|_{\omega=\omega_0}\beta
\end{equation}
\vspace{-2mm}

\noindent and subsequently the velocity of center of curvature will be

\vspace{-2mm}
\begin{equation}
    \dot{\vec{o}}=-v_{R_0}(\cos\gamma\, \hat{x} + \sin\gamma\,\hat{y})
\end{equation}
\vspace{-3mm}

\noindent To move $\vec{o}$ towards a certain target position $\vec{o}_g$, a simple solution can be

\vspace{-2mm}
\begin{equation}
    v_R = - k_p (\vec{o}_g-\vec{o})\cdot(\cos\gamma\, \hat{x} + \sin\gamma\,\hat{y})
    \label{eqn:simplecontrol}
\end{equation}
\vspace{-3mm}

\noindent where $k_p > 0$ is a constant. It is worth noting that in experiments, $\vec{o}$ will not be the same as the actual center of curvature due to the transient behavior of Rollbot.

\begin{figure}[!h]
\centering
\vspace{-2mm}
\includegraphics[width=\linewidth]{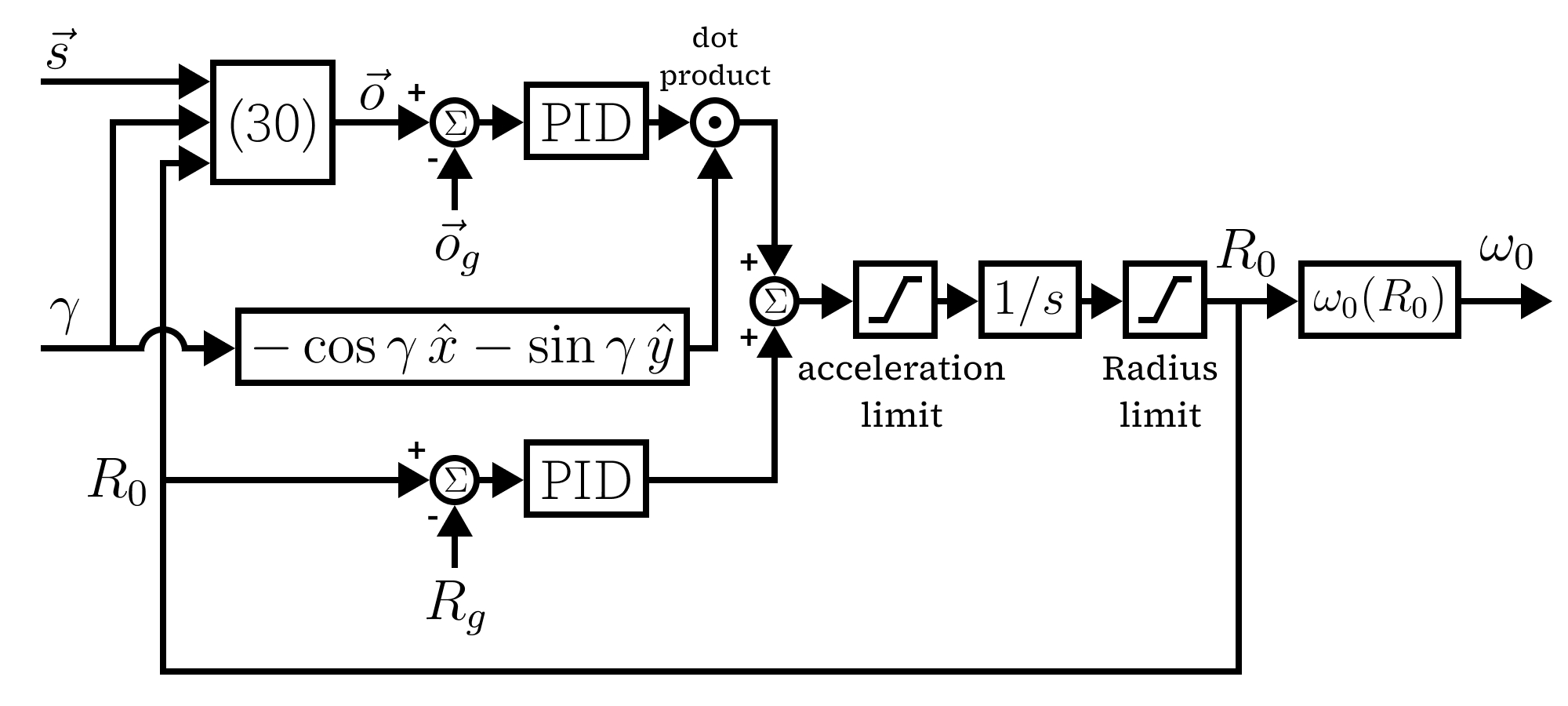}\\
\vspace{-3mm}
\caption{Control diagram used in experiments. The input of the system are the current position $\vec{s}$ and orientation $\gamma$ of the shell provided by the Optitrack system and the output is the target driving speed $\omega_0$.}
\vspace{-1mm}
\label{fig:controldiag}
\end{figure}

In reality, we need to consider the capability of the motor and various non-idealities. So, instead of using (\ref{eqn:simplecontrol}), we add a limit to the revolving radius to limit the driving speed, a limit to the maximum acceleration so Rollbot is always near quasi-stable state, and use PID instead of simple proportional control to compensate for external interference like the slope of the ground and to keep the system stable. The control diagram is shown in Fig. \ref{fig:controldiag}.

So far we have been talking about how to control $\vec{o}$ instead of $\vec{s}$. In case we want the robot to move across a certain point $\vec{s}_g$ at velocity $\dot{\vec{s}}_g$, we only need to compute the corresponding target $R_g$ using equations obtained in section \ref{sec:static} and target $\vec{o}_g$ using (\ref{eqn:ofunc}), and command Rollbot to roll around $\vec{o}_g$ at revolving radius $R_g$. If we want to stop the Rollbot at $\vec{s}_g$, we will first let it revolve around $\vec{o}_g$ and gradually stop the motor when Rollbot is approaching $\vec{s}_g$.

\section{Experiments}\label{sec:experiments}

We conducted several experiments to demonstrate the ability of Rollbot. We used Optitrack system to provide Rollbot with its real-time position and orientation, but all other computations are executed on-board.

\subsection{Open-loop Motion}\label{sec:open}

\begin{figure}[htbp]
\centering
\vspace{-2.5mm}
\includegraphics[width=\linewidth]{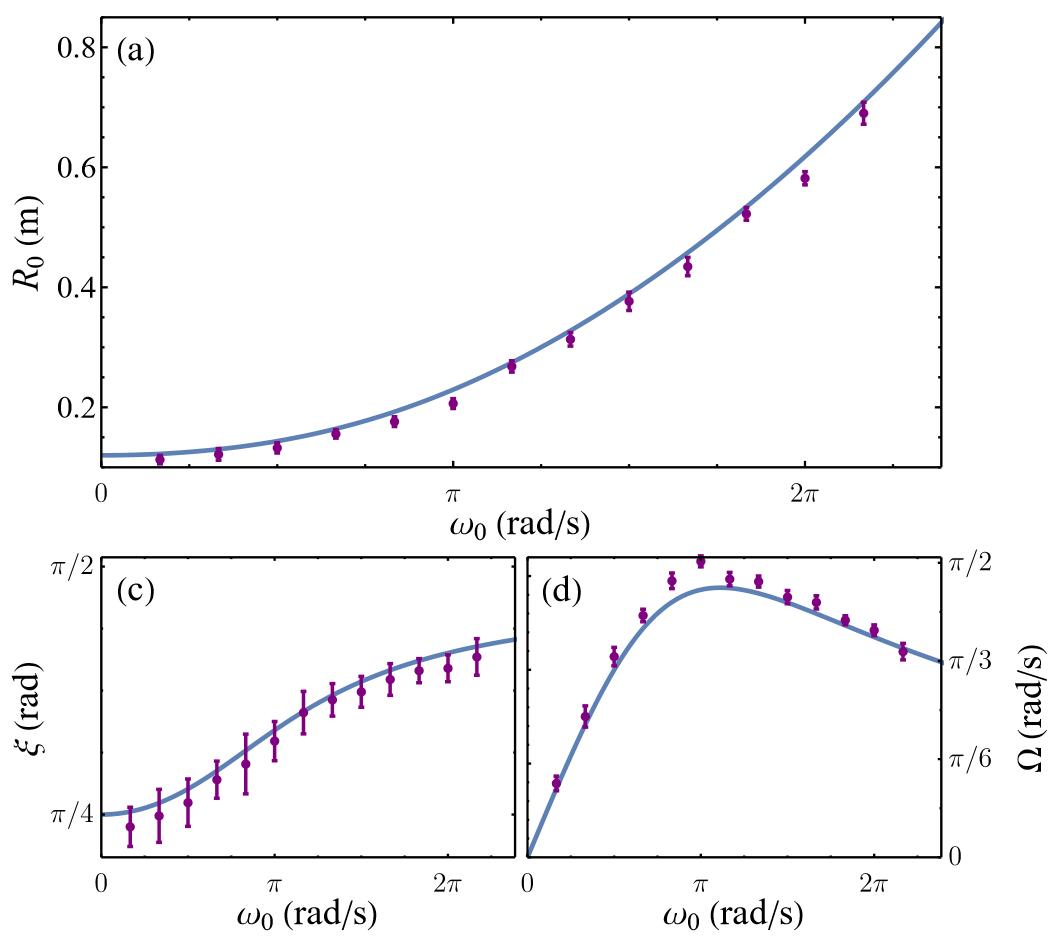}\\
\vspace{-3mm}
\caption{Comparison between the theoretical and experimental results. Theoretical curves are drawn in blue lines and experimental results are drawn in purple points. The slight systematic difference between experimental data and theoretical prediction of $R_0$ and $\xi$ (the difference is $6\,$mm and $1^\circ$ respectively) are likely caused by the slight imbalance of the shell. The exact values of $\xi$ were not measured since the pendulum is inside the opaque sphere. However, we confirmed qualitatively that it always remain near the bottom of the shell during operation.}
\vspace{-3.5mm}
\label{fig:comparison}
\end{figure}

In the open-loop motion experiments, Rollbot will maintain a constant driving speed from $0$ to $7\,$rad/s. We collect the revolving radius $R_0$, the rotating axis' tilting angle $\xi$, and the revolving angular velocity $\Omega$ of the Rollbot under different driving speeds $\omega_0$ and compare it with the theoretical result obtained in section \ref{sec:static} to verify that our theory and approximation are consistent with reality. The results are shown in Fig. \ref{fig:comparison}.

\subsection{Stable Circular Motion}\label{sec:circular}

\begin{figure}[htbp]
\centering
\includegraphics[width=0.85\linewidth]{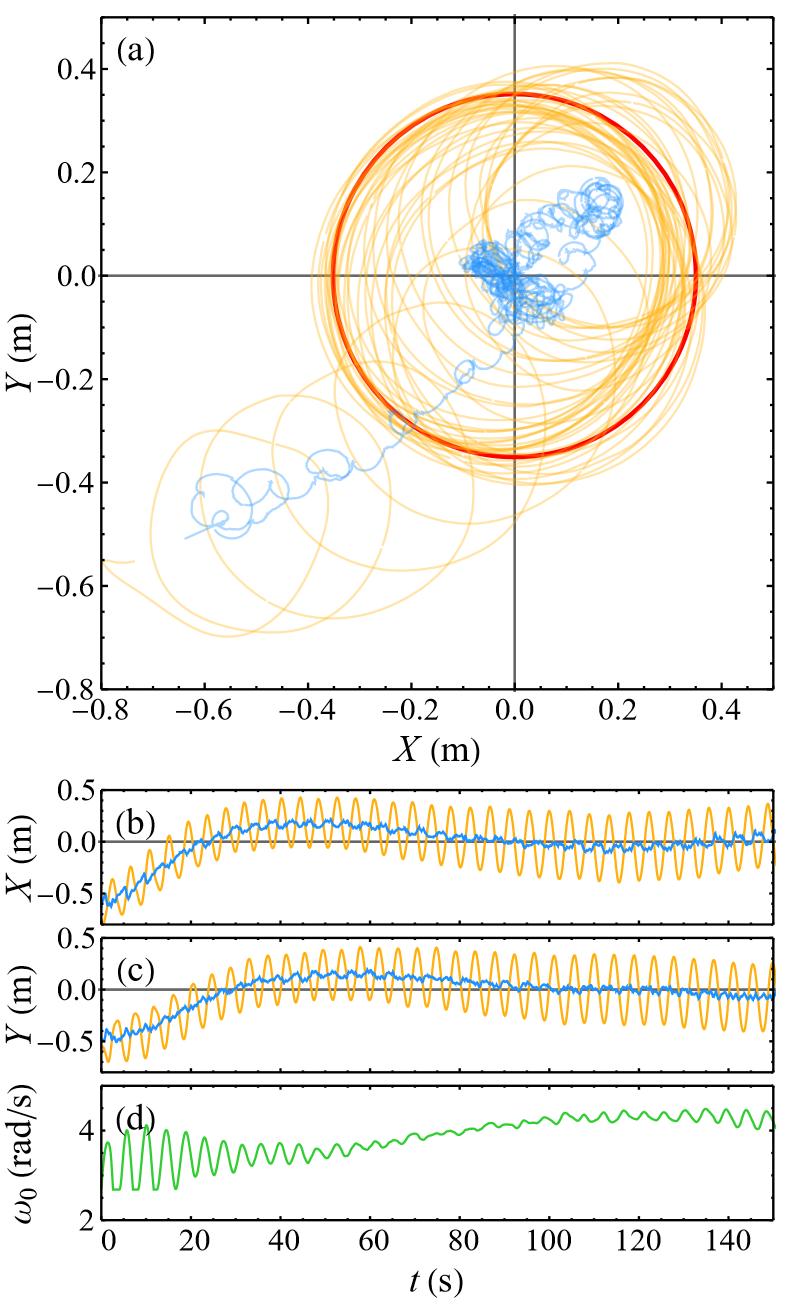}\\
\vspace{-3.5mm}
\caption{Trajectory of Rollbot when moving towards the set point. (a) shows the trajectory, where the red circle shows the target radius $R_g=0.35\,$m, and the orange and blue lines are the trajectory of Rollbot and the $\vec{o}$ respectively. (b), (c), and (d) shows the change of variables over time, where the orange and blue lines represent the position of Rollbot and the $\vec{o}$ respectively and the green line represent the driving speed.}
\vspace{-1mm}
\label{fig:move}
\end{figure}

\begin{figure}[!t]
\centering
\includegraphics[width=.95\linewidth]{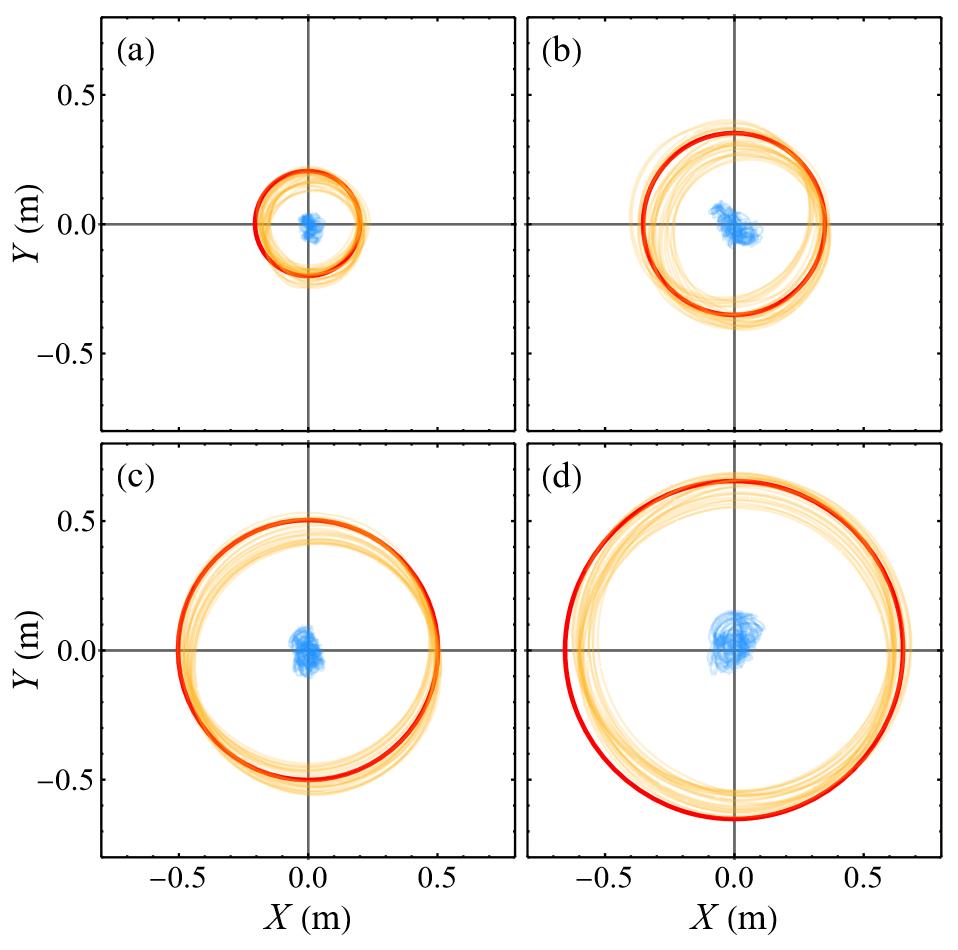}\\
\vspace{-3mm}
\caption{Trajectory of Rollbot when stably revolving around the set point at \{0,0\}. Target revolving radius $R_g$ is $0.20,0.35,0.50,0.65\,$m for (a) to (d) respectively. \edit{}{The ground tilts slightly upward toward the top left.}}
\vspace{-3mm}
\label{fig:revolve}
\end{figure}

In the stable circular motion experiments, Rollbot will try to circle around a set point with a given radius ranging from $20\,$cm to $65\,$cm. Fig. \ref{fig:move} shows how Rollbot can start from a point far away from the setpoint and move towards the goal, and Fig. \ref{fig:revolve} shows the trajectory of Rollbot when it is stably revolving around the set point. It is worth noting that the ground in our experiments was tilted by approximately $1^\circ$, which will cause Rollbot to drift rapidly from the target trajectories without feedback control. With the proposed feedback policy, however, Rollbot can successfully compensate for this bias.

From Fig. \ref{fig:move} we can see that Rollbot moves towards the set point by having higher driving speed when moving towards the set point and vice versa. In this example, Rollbot moves at an average speed of approximately $2\,$cm/s when moving towards the set point. Once it reached the stably revolving state, it will still modulate the driving speed slightly to compensate for the slope of the ground. 

\edit{}{As shown in Fig. \ref{fig:revolve}, Rollbot can revolve around the set point at the specified radius. During these tests, Rollbot is within 15\,cm away from the expected trajectory, with an average deviation of 3.4\,cm.}

\subsection{Moving between Waypoints}\label{sec:path-following}

\begin{figure}[!t]
\centering
\includegraphics[width=.95\linewidth]{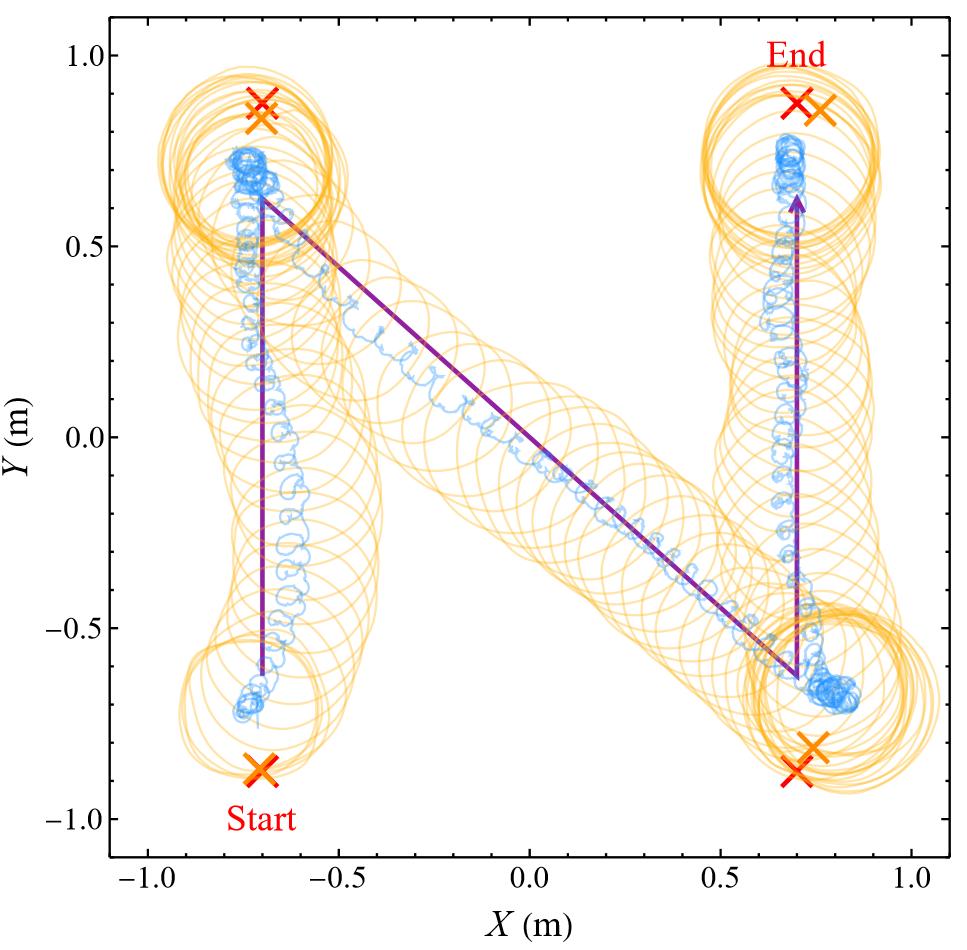}\\
\vspace{-3mm}
\caption{Experimental results for waypoint movement experiment. Purple line shows the target trajectory of $\vec{o}$, red and orange crosses are the target and actual stopping points.}
\vspace{-5mm}
\label{pathmove}
\end{figure}

We would also like to demonstrate the Rollbot's ability to maneuver between waypoints, so we programmed Rollbot to move in a `N' shape and stop at the four vertices. The results in Fig. \ref{pathmove} shows that Rollbot can successfully move towards waypoints and stop within $7\,$cm of the targets.

\section{Discussion}\label{sec:discussion}

The presented Rollbot prototype is primarily a proof-of-concept, focusing on demonstrating that precise planar locomotion is possible with a single actuator. In our experiments, we used an OptiTrack system for high-speed (100Hz sampling rate) global localization to simplify testing; in the future, we plan to switch to IMU-based sensing for fully onboard control. At present, Rollbot travels at about $2\,$ cm/s, much slower than typical rolling robots, which can reach $10\,$cm/s to $1\,$m/s. This is largely due to our quasi-stable motion assumption and the resulting conservative control strategy. Future work will focus on drastically improving performance through better parameter tuning and advanced controllers that can take advantage of the robot’s full dynamic capabilities. Another promising direction is adapting Rollbot for more challenging environments, such as sand, gravel, or even water surfaces, which will further expand the range of scenarios where single-actuator spherical robots can be useful. \edit{}{Aside from these possible movement enhancements, sensing and telemetry can also be equipped on Rollbot, allowing it to execute a wider range of tasks autonomously.} With these improvements, Rollbot can become \edit{the simplest}{a minimalist} yet still powerful spherical robot, making it a great testbed for underactuated robotics and related fields.

\edit{}{Comparing with the other single-actuator spherical robot proposed in \cite{spigler2024intuitive}, Rollbot differs in two fundamental aspects. First, Rollbot has a more symmetric mass distribution, which significantly reduces the complexity of its dynamics. This simplicity allows for the derivation of closed-form equations of its quasi-stable state, which serve as the foundation for the described control law. Second, while the mechanism in \cite{spigler2024intuitive} was demonstrated primarily in simulation, Rollbot is designed for operation in real world, which can greatly benefit from robustness. By operating within a quasi-stable regime, Rollbot trades peak locomotion speed for increased stability and resilience against environmental perturbations, such as surface friction variations and minor ground tilt.}

\section{Conclusion}\label{sec:conclusion}

In this paper, we presented Rollbot, a single actuator spherical robot capable of controlled rolling motion on the ground. We provided a general mechanical analysis of spherical robots driven by internal masses and the quasi-stable state and stability analysis of Rollbot. Based on our theory, we designed Rollbot's hardware and control algorithm and successfully achieved stable circular motion and waypoint following. Through experiments, we also verified our theory's accuracy.

We hope the design and analysis of Rollbot will inspire new minimalist robot designs, and the proposed control strategy can be used as a fail-safe for conventional spherical robots when they experience partial motor failure.

\printbibliography

\end{document}